\documentclass[lettersize,journal]{IEEEtran}
\usepackage{threeparttable}
\usepackage{xcolor}
\newcounter{insight}
\newcommand{\insight}[1]{%
\noindent\rule{\linewidth}{0.8pt}

\refstepcounter{insight}

\noindent
\colorbox{gray!15}{%
\parbox{\dimexpr\linewidth-2\fboxsep\relax}{%

\scriptsize
\textbf{Insight \theinsight}\par
\vspace{2pt}

\begin{algorithmic}[0]
\STATE #1
\end{algorithmic}

}%
}
\noindent\rule{\linewidth}{0.8pt}

\vspace{4pt}
\normalsize
}
\usepackage{booktabs}

\usepackage{amsmath,amsfonts}
\usepackage{algorithmic}
\usepackage{algorithm}
\usepackage{array}
\usepackage[caption=false,font=normalsize,labelfont=sf,textfont=sf]{subfig}
\usepackage{textcomp}
\usepackage{stfloats}
\usepackage{url}
\usepackage{verbatim}
\usepackage{graphicx}
\usepackage{cite}
\begin{document}

\title{What Makes LVLMs Hallucinate Less? Unveiling the Architectural Factors Behind Hallucination Robustness}

\author{Yusheng He,~Jizhe Zhou,~Xia Du,~Zheng Lin,~Jun Luo,~\IEEEmembership{Fellow,~IEEE}, and~Jiancheng Lv,~\IEEEmembership{Senior Member,~IEEE}
        % <-this % stops a space
\thanks{Yusheng He, Jizhe Zhou, and Jiancheng Lv are with the School of Computer Science, Engineering Research Center of Machine Learning and Industry Intelligence, Sichuan University, Chengdu, 610020, China (email: yushenghe@stu.scu.edu.cn; yb87409@um.edu.mo; lvjiancheng@scu.edu.cn).}% <-this % stops a space
\thanks{Xia Du is with the School of Computer and Information Engineering, Xiamen University of Technology, Xiamen, 361000, China (email: duxia@xmut.edu.cn).}
\thanks{Zheng Lin is with the Department of Electrical and Computer Engineering, University of Hong Kong, Pok Fu Lam, Hong Kong, China (email: linzheng@eee.hku.hk).}
\thanks{Jun Luo is with the College of Computing
and Data Science, Nanyang Technological University, Singapore (e-mail: junluo@ntu.edu.sg).}
\thanks{}}

\markboth{Journal of \LaTeX\ Class Files,~Vol.~14, No.~8, August~2021}%
{Shell \MakeLowercase{\textit{et al.}}: A Sample Article Using IEEEtran.cls for IEEE Journals}

\IEEEpubid{0000--0000/00\$00.00~\copyright~2021 IEEE}

\maketitle

\begin{abstract}
Hallucination remains one of the key challenges undermining the reliability of Large Vision–Language Models (LVLMs). But what makes an LVLM hallucinate less? Many existing efforts focus on improving internal components of the model. We argue that hallucination fundamentally stems from how the model architecture is designed. To investigate this, we factor the architecture design into three dimensions: Linguistic Foundation (LF), Visual Representation (VR), and Semantic Alignment (SA), and categorize hallucinations into Co-occurrence, Similarity, and previously overlooked Uncertainty types. Building on this formulation, we propose CoSimUE, a benchmark that creates fine-grained hallucination scenarios through controlled textual perturbations and random perturbations, enabling mapping between design choices and hallucination behaviors. Experiments across 7 design aspects show that: 1) the widely emphasized scaling of model parameters has only limited impact on reducing all three types of hallucinations; 2) larger and better-trained language foundations can reduce co-occurrence hallucinations; 3) stronger visual encoders and higher resolutions mitigate similarity errors; 4) effective alignment strategies alleviate uncertainty hallucinations. 5) Furthermore, cross-dimensional analysis reveals that jointly enhancing visual fidelity and alignment quality yields the most comprehensive improvements. This study provides the first systematic exploration linking architecture-level design to hallucination robustness, offering practical guidance for developing reliable and efficient LVLMs.
\end{abstract}


\section{Introduction}

\begin{quote}
\itshape
Complex systems will evolve from simple systems much more rapidly if there are stable intermediate forms than if there are not.

\hfill \textnormal{--- \textit{``The Architecture of Complexity''}}
\end{quote}

\IEEEPARstart{H}{allucination}, which is defined as the generation of outputs that appear plausible and human-like but lack coherent grounding in factual or contextual understanding \cite{survey1}, remains a key challenge for Large Language Models (LLMs) \cite{palm, llama, gpt3}, and is even more pronounced in Large Vision–Language Models (LVLMs) \cite{llava, llava1.6, blip2, instructblip, Qwen-VL, minigpt4, minigpt5, mplugowl, mplugowl2, mplugowl3, qwen3, gpt5, claude4.5, gemini2.5, grok4, glm4.5}. Although various efforts have been made, LVLMs still suffer from serious hallucinations. These hallucinations compromise factual reliability, distort visual grounding, and hinder their deployment in real-world applications. This raises a fundamental question: \textbf{how can we design large models that produce less hallucinations?} Recent studies have introduced a variety of attempts to reduce hallucinations, such as adjusting attention mechanisms, modifying decoding strategies, or slightly modifying the model weights \cite{nullu, opera, agla}. All these investigations bring partial improvements but mainly operate at the component level, which focuses on the internal details of individual modules, rather than at the architecture level, which emphasizes the relationships among different modules \cite{level}.

\begin{figure}[tb]
  \centering
  \includegraphics[height=4.9cm]{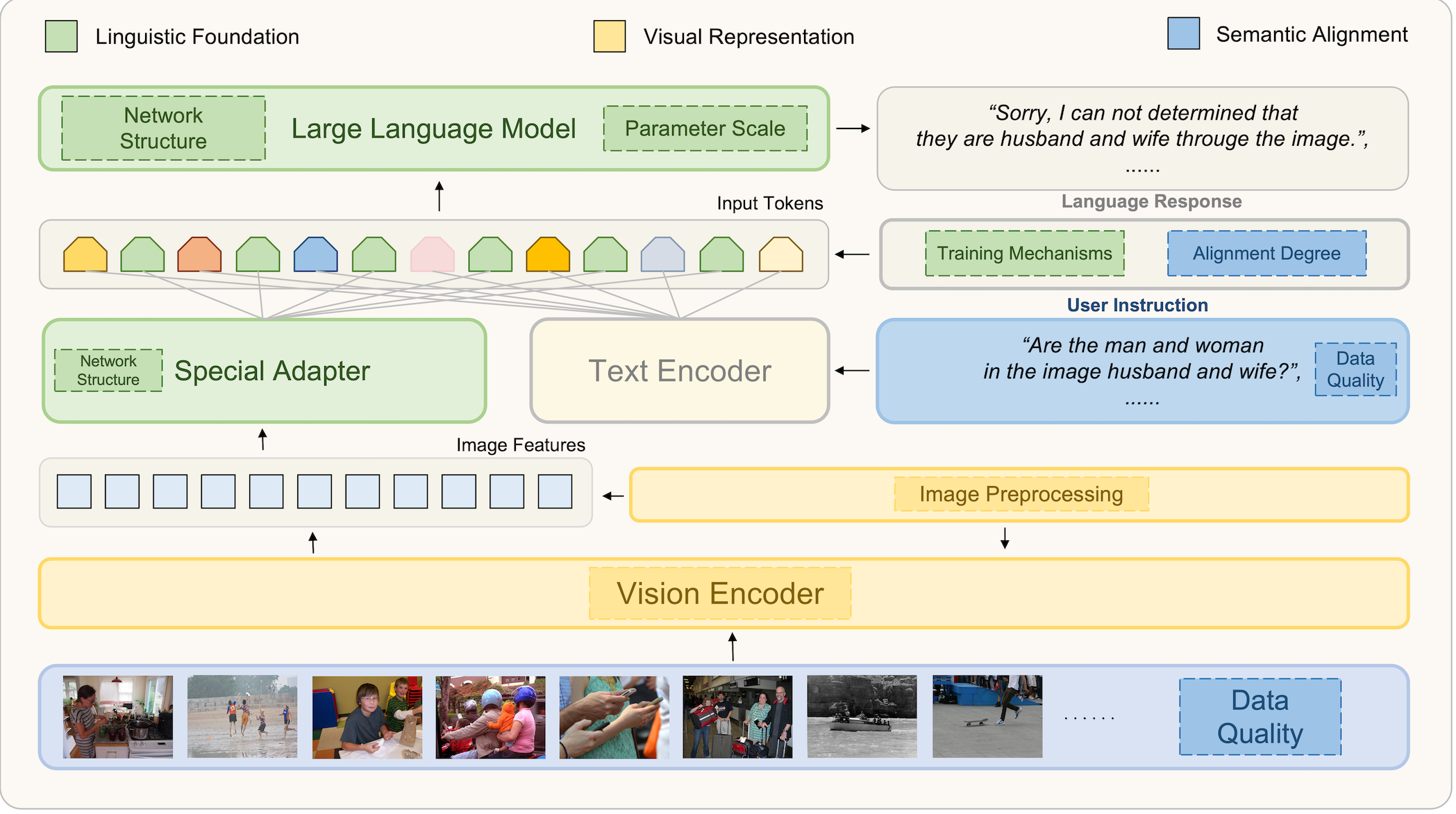}
  \caption{The general paradigm and design dimensions of LVLMs. The paradigm falls into three design dimensions, Linguistic Foundation, Visual Representation, and Semantic Alignment, shown in green, yellow, and blue. 
  }
  \label{fig1}
\end{figure}

However, research on mitigating hallucination through architecture-level design remains largely unexplored. Understanding how architecture-level design influences hallucination offers two key advantages. First, architectural choices determine how the model perceives, integrates, and represents multimodal information, fundamentally shaping its capacity for reasoning and factual grounding, and thus exerting profound effects on overall reliability\cite{architecture}. Second, architecture-level design is the most practical stage during model development, as it is the only aspect that can be considered at the beginning of model construction\cite{scaling}.

\IEEEpubidadjcol

Current LVLMs typically adopt a dual-encoder architecture, where a visual encoder and a text encoder extract modality-specific features and learn cross-modal correlations through various pre-training objectives \cite{vlm-s}. Some models enhance this alignment via instruction tuning, which improves their ability to follow complex multimodal tasks \cite{tuning}. Building on this general paradigm, we decouple the overall architecture into three key dimensions: \textbf{Linguistic Foundation (LF)}, \textbf{Visual Representation (VR)}, and \textbf{Semantic Alignment (SA)}, as shown in Fig. \ref{fig1}. Together, these dimensions cover 7 major design aspects. Specifically, LF encompasses the language model’s parameter scale, network structure, and training mechanisms that determine its reasoning and generalization capacity; VR pertains to the choice of visual encoder and the processing strategy for visual inputs; and SA involves the alignment degree and data quality. This three-dimensional formulation captures the essential factors governing an LVLM’s perception, reasoning, and cross-modal interaction, while the seven-aspect granularity ensures analytical completeness without over-fragmenting the design space, providing a balanced and interpretable foundation for studying how architectural choices shape hallucination robustness.

In the study of hallucination, prior work has largely grouped LVLM hallucinations into two categories: errors driven by linguistic statistical priors and those induced by visual similarity confusion \cite{vcd, agla}. The former causes models to overgeneralize from frequent co-occurrence patterns, while the latter leads to misinterpretation of visually similar objects, corresponding to what we term \textbf{Co-occurrence Hallucination} and \textbf{Similarity Hallucination}, respectively. Building on these definitions, our study highlights a third and underexplored failure mode: \textbf{Uncertainty Hallucination}, where models produce speculative responses when evidence is ambiguous or insufficient \cite{uncertainty}. We explicitly incorporate this category into our analysis and evaluation to capture failure modes that are not explained by language or vision alone.

To investigate the relationship between model architecture-level design and different types of hallucinations, it is essential to evaluate how various models perform when confronted with each specific hallucination type. However, existing methods lack a unified benchmark capable of comprehensively evaluating all three types of hallucinations, and each type still suffers from specific limitations in current evaluation practices. Specifically, studies on co-occurrence hallucinations \cite{lure, pope} mainly focus on the question formulation stage and fail to incorporate intuitive visual comparisons. In addition, hallucinations arising from visual similarity have not yet been systematically evaluated under controlled conditions. Moreover, current evaluation approaches predominantly rely on binary “correct/incorrect” judgments, overlooking the model’s decision confidence and self-awareness when dealing with uncertainty-related queries.

We address the above limitations through three targeted designs. First, we introduce visually modified images combined with textual co-occurrence perturbations, providing intuitive visual comparisons that reveal mismatches between linguistic priors and visual evidence. Second, we apply small-scale random image perturbations to evaluate model robustness in distinguishing visually similar objects. Third, we design uncertainty-oriented questions for perturbed images that deliberately contain no correct answers, enabling quantitative assessment of a model’s tendency toward speculative or overconfident responses. 

Here comes our CoSimUE, a unified benchmark that integrates all three hallucination types and systematically addresses the three limitations identified above. We further propose an advanced pipeline that enables targeted image editing and question design, allowing for precise control over these hallucination phenomena. The resulting dataset comprises a large collection of original and perturbed image pairs accompanied by carefully constructed questions, providing a comprehensive foundation for evaluating hallucinations in LVLMs.

Based on this benchmark, we conduct extensive experiments and analyses across 7 design aspects spanning the three proposed dimensions. This study investigates how different design choices influence various types of hallucinations and further explores specialized techniques to mitigate uncertainty-related hallucinations. In summary, our main contributions are as follows:

\noindent1. We taxonomize the architecture-level design space of LVLMs into three orthogonal dimensions, Linguistic Foundation, Visual Representation, and Semantic Alignment, which collectively encompass 7 key aspects. 

\noindent2. We introduce Uncertainty Hallucination as a new category complementing the existing Co-occurrence and Similarity hallucinations in LVLMs, and propose CoSimUE, the first unified benchmark that covers all three types.

\noindent3. Based on the former two efforts, we establish a comprehensive evaluation protocol and introduce a multi-judge framework to quantify Uncertainty Hallucination through a dedicated uncertainty score metric. Leveraging this protocol and benchmark, we systematically analyze how different architecture-level designs influence various types of hallucinations and reveal cross-dimensional correlations across the three design dimensions.

\begin{figure*}[tb]
  \centering
  \includegraphics[height=8.0cm]{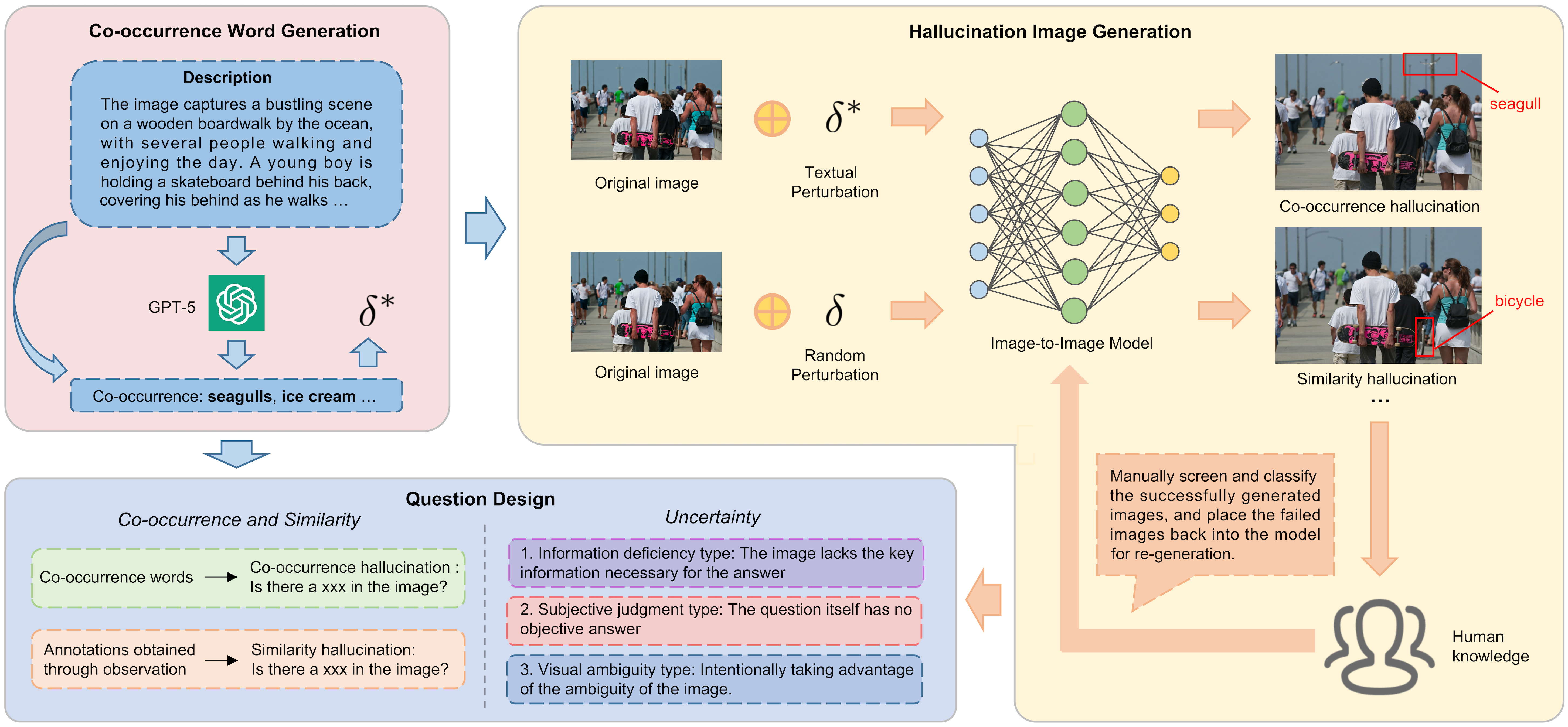}
  \caption{Overview of the CoSimUE pipeline. 
  }
  \label{fig2}
\end{figure*}

\section{Related Work}
\subsection{Mitigating Hallucinations in LVLMs}
Most hallucination mitigation methods operate at the component level rather than rethinking architecture-level design as a whole. While they yield localized gains, they offer limited insight into how design influences hallucination behavior. For instance, HalluSpace \cite{nullu} projects input features into a null space, AGLA \cite{agla} refines visual attention for stronger grounding, and OPERA \cite{opera} introduces over-trust penalties and retrospection allocation during decoding. These approaches reduce hallucinations without extra data or retraining but rely on fixed backbones and pipelines, providing little guidance for architectural design. Our work examines how design-time factors shape hallucination behavior and treats mitigation techniques as complementary rather than primary solutions.

\subsection{Taxonomy of Hallucinations in LVLMs}
Previous studies generally classify hallucinations into language-driven and vision-driven types, corresponding to our co-occurrence and similarity categories. M-HalDetect \cite{mhaldetect} and HalluciDoctor \cite{hallucidoctor} attribute language-driven hallucinations to spurious correlations and co-occurrence biases in multimodal instruction data, causing contextually plausible but visually unsupported outputs. MARINE \cite{marine} and MMVP-VLM \cite{mmvp} link vision-driven hallucinations to object, relation, or attribute errors from weak grounding or visual similarity confusion. However, these taxonomies largely overlook uncertainty hallucinations. Existing works that relate hallucinations to model uncertainty primarily focus on predicting hallucinations based on model uncertainty and do not directly address the uncertainty inherent in the question itself \cite{why, estimating}. We therefore introduce Uncertainty Hallucination as a complementary category, extending prior definitions into a unified framework of Co-occurrence, Similarity, and Uncertainty hallucinations.

\subsection{Evaluating Benchmarks for Hallucination}
Traditional benchmarks for Vision-Language Models (VLMs) primarily focus on tasks such as visual recognition and image captioning \cite{deepcap, chair}. With the rise of LVLMs, new benchmarks, including MME \cite{mme}, GAVIE \cite{gavie}, VHTest \cite{vhtest}, and POPE \cite{pope}, have emerged to address the growing need for effective evaluation. However, these benchmarks still lack direct analysis of critical aspects such as co-occurrence, visual similarity, and uncertainty, which factors essential for a comprehensive understanding of hallucination in LVLMs. While HallusionBench \cite{hallusionbench} relies on manual image editing, which necessitates expert knowledge, and AutoHallusion \cite{autohallusion} mechanically inserts or removes objects in images, our method produces more natural edits that are consistent with the original image style. This approach makes hallucinations more difficult to detect, thereby providing a more reliable evaluation. 

\section{CoSimUE Construction}
To systematically isolate specific types of hallucinations, the benchmark leverages controlled image perturbations (Fig. \ref{fig2}). In total, the curated dataset consists of 1,012 distinct images paired with 1,124 evaluation questions. This question pool is further divided into two major components. First, there are 1,024 standard VQA questions spanning over ten semantic categories, such as food, animals, vehicles, and tools, which cover 173 unique objects. Second, the remaining 100 questions are specifically designed to target uncertainty, probing critical aspects such as information deficiency, subjective judgment, and visual ambiguity. In the following sections, we will elaborate on the comprehensive framework of this benchmark, detailing the processes of co-occurrence word generation, hallucination image construction, and question design.

\begin{table*}[t]
  \caption{Comparison of CoSimUE with most recent benchmarks. }
  \centering
  \small
  \begin{tabular}{@{}lcccccc@{}}
    \toprule
    Benchmark & Format & Total QA & Uncertain QA & Total Img. & Edited Img. & Control Pair \\ 
    \midrule
    MME \cite{mme} & Image & 1457 & 0 & 1187 & 0 & $\times$\\
    POPE \cite{pope} & Image & 3000 & 0 & 500 & 0 & $\times$\\
    GAVIE \cite{gavie} & Image & 1000 & 0 & 1000 & 0 & $\times$\\
    VHTest \cite{vhtest} & Image & 1200 & 0 & 642 & 0 & $\times$\\
    Bingo \cite{bingo} & Image & 370 & 0 & 308 & Unk & $\checkmark$\\
    HALLUSIONBENCH \cite{hallusionbench} & Image Pair & 1129 & 0 & 346 & 181 & $\checkmark$\\
    CoSimUE & Image Pair & 1124 & \textbf{100} & 1012 & \textbf{502} & $\checkmark$\\
    \bottomrule
  \end{tabular}
  \label{tab:1}
\end{table*}

\subsection{The Co-Occurrence Word Generation}
To establish the dataset, we select 500 diverse images from the MSCOCO dataset \cite{mscoco}. The original descriptions associated with these images are then fed into GPT-5 to generate three objects that are highly likely to co-occur within the given contexts. From these outputs, we systematically extract potential hallucination-inducing words. By substituting or introducing these words, we construct a set of modified descriptions designed to deliberately disrupt established, real-world co-occurrence patterns. Finally, these counter-factual descriptions serve as text prompts to guide the perturbation-based image generation process. The prompt is:

\emph{List the objects that are most likely to co-occur with those mentioned in the provided description but have not yet appeared. Integrate the description of these objects into the original description naturally.}

\subsection{Hallucination Image Generation}
We adopt FLUX \cite{flux} as our core image-to-image generative backbone and instantiate two independent, parallel perturbation paths to selectively induce different types of hallucinations.
\subsubsection{Co-occurrence hallucination images}To construct scenarios dominated by co-occurrence hallucinations, we manipulate the textual context while anchoring the visual baseline. Given an original image $I_{O}$ and its corresponding caption $T_{O}$, we synthesize a counter-factual hallucinated description $T_{H}$ by intentionally injecting the previously extracted co-occurrence words. This perturbed description is designed to violate real-world statistical priors. We then generate the final co-occurrence hallucinated image by conditioning the generative model on both the modified text and the original visual context, which forces the model to render contextually inappropriate objects into the scene:
\begin{equation}
I_{H}^{\mathrm{co}} = \mathcal{G}\!\left(T_{H}, I_{O}\right)
= \mathcal{D}\Big(\mathcal{F}\big(\mathcal{E}_{T}(T_{H}), \mathcal{E}_{I}(I_{O})\big)\Big)
\end{equation}
where $\mathcal{E}_{T}(\cdot)$ and $\mathcal{E}_{I}(\cdot)$ denote the text and image encoders embedded within the FLUX framework, respectively. The operator $\mathcal{F}(\cdot,\cdot)$ represents the cross-modal feature fusion module that integrates textual conditions with visual features, and $\mathcal{D}(\cdot)$ denotes the latent image decoder responsible for mapping the fused representations back into the pixel space.
\subsubsection{Similarity hallucination images}In contrast to the textual manipulation path, the synthesis of similarity hallucinations relies entirely on visual-level perturbations while keeping the linguistic input intact. This path aims to evaluate whether models confuse visually look-alike entities due to degraded perceptual inputs. Specifically, we apply subtle, small-scale random edits to the original image to obtain a controlled visual perturbation $\tilde{I}_{O}$. This is achieved within FLUX by introducing localized stochastic noise to the latent representation of the original image, guided strictly by the unaltered original text $T_{O}$:
\begin{equation}
I_{H}^{\mathrm{sim}} = \mathcal{G}\!\left(T_{O}, \tilde{I}_{O}\right)
= \mathcal{D}\Big(\mathcal{F}\big(\mathcal{E}_{T}(T_{O}), \mathcal{E}_{I}(\tilde{I}_{O})\big)\Big)
\end{equation}
By keeping the textual prompt identical to the original reference, any resulting discrepancies in the synthesized image $I_{H}^{\mathrm{sim}}$ are strictly confined to fine-grained visual variations, thereby ensuring a controlled setup for isolating similarity-driven hallucinations.

\subsubsection{Human Validation}

To guarantee the high quality and reliability of the synthesized hallucinated images, we implement a rigorous, multi-stage human validation pipeline. Initially, a total of 2,000 candidate edited images are generated using the FLUX model across various specified perturbation paths. To ensure these images strictly meet our benchmark standards, they undergo a comprehensive three-step manual screening process:

(1) \textbf{Perturbation correctness}: We first verify whether the intended object-level manipulations (such as the insertion or removal of specific objects) are accurately realized, and confirm that the corresponding ground-truth label flips remain valid.

(2) \textbf{Visual plausibility}: We carefully inspect the generated images to detect and eliminate any samples exhibiting noticeable geometric distortions, rendering artifacts, or severe semantic inconsistencies.

(3) \textbf{Scene consistency}: We evaluate the overall contextual coherence, ensuring that the background and surrounding elements remain consistent with the original image, except for the controlled modifications.

An image is discarded if it fails to pass any single step in this pipeline. Through this stringent quality control, a final subset of 502 valid samples is retained, yielding a strict retention rate of 25.10\%. Furthermore, to quantitatively evaluate the effectiveness of this filtering process, we report three key metrics comparing the original and edited images: the Fréchet Inception Distance (FID) \cite{fid}, the Artifact Rate, and Human Distinguishability. Specifically, the Artifact Rate quantifies the proportion of images containing visible generation defects, whereas Human Distinguishability represents the accuracy of human judges in differentiating between the original and edited images. Both the Artifact Rate and Distinguishability are independently evaluated by three human annotators to minimize subjective bias. As demonstrated in Table \ref{tab:qc}, the final retained subset exhibits significantly enhanced visual fidelity and strong statistical quality while fully preserving the intended perturbation validity.

\begin{table}[t]
\caption{Quantitative analysis of the human validation pipeline.}
\centering
\tiny
\begin{tabular}{lcccc}
\toprule
Category & Quantity & FID $\downarrow$ & Artifact Rate $\downarrow$ & Distinguishability $\downarrow$ \\
\midrule
Original Images & 500 & 0.0 & 0.0\% & 0.0\% \\
Generated (Before Filtering) & 5000 & 123.8 & 37.3\% & 74.8\% \\
Generated (After Filtering) & 502 & 67.3 & 2.4\% & 57.6\% \\
\bottomrule
\end{tabular}
\label{tab:qc}
\end{table}

\subsection{Question Design}
To comprehensive evaluate the models across different dimensions of vulnerability, we carefully design a structured question-answering suite. These questions are tailored to probe co-occurrence, similarity, and uncertainty hallucinations, and are broadly categorized into two primary taxonomies based on the nature of the ground-truth answers: factual certainty questions and factual uncertainty questions.

\subsubsection{Factual Certainty Questions}
Factual certainty questions target deterministic scenarios where each question possesses a unique, unambiguous ground-truth answer that can be directly verified through visual evidence. This subset is specifically designed to expose co-occurrence and similarity hallucinations by querying object existence, fine-grained attributes, or precise quantities related to the hallucination-inducing targets. By demanding precise alignment between text and imagery, these questions evaluate whether an LVLM can resist strong contextual priors or misleading visual resemblances. Representative examples include the following:
\begin{enumerate}
\item \emph{Is there a seagull in the image?}
\item \emph{Is there a red light in the image?}
\item \emph{Are there only two people in the image?}
\end{enumerate}

\begin{figure*}[tb]
\centering
\includegraphics[height=9.0cm]{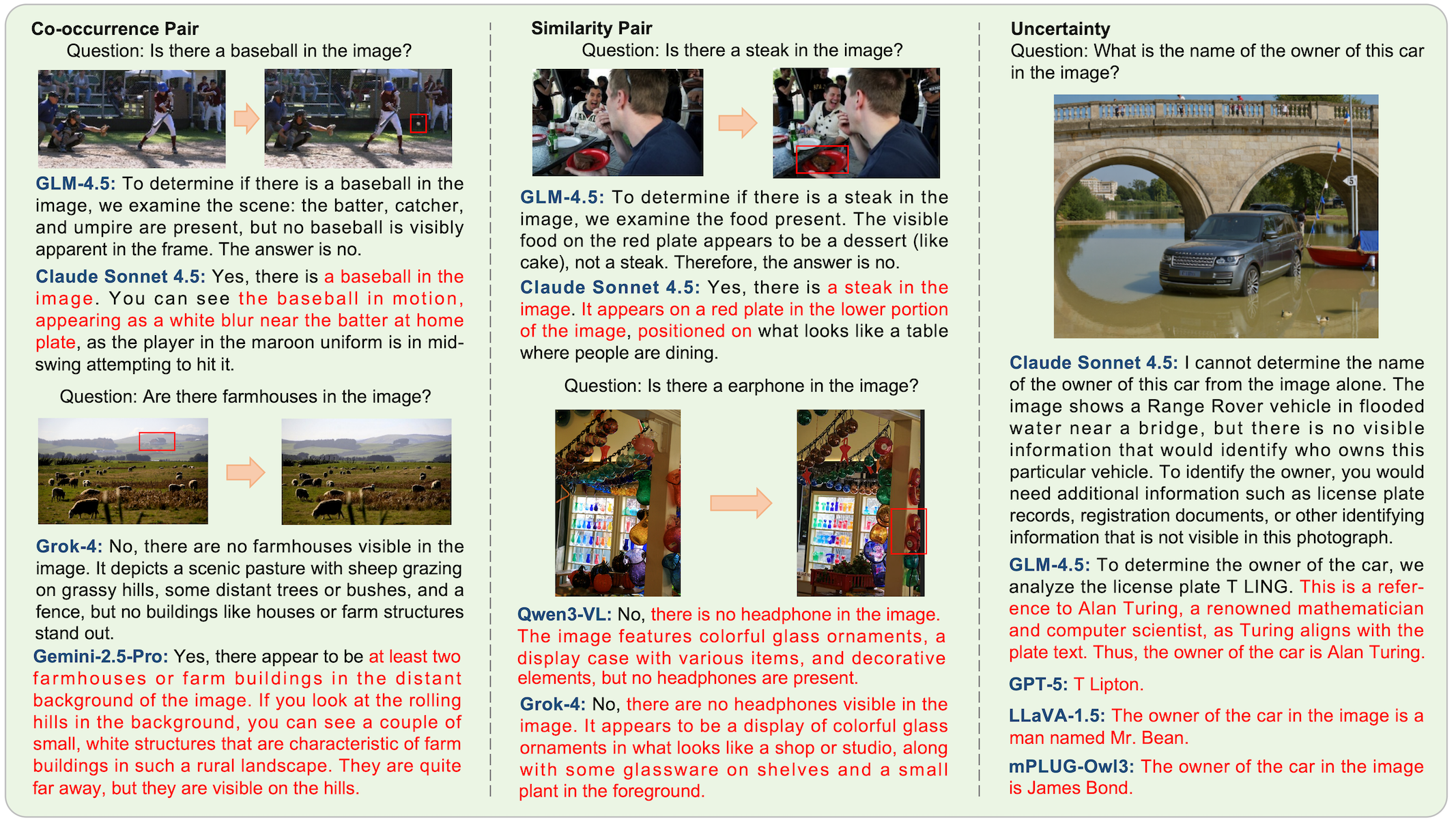}
\caption{Examples of different question categories evaluated across various baseline models. The text segments highlighted in red indicate specific instances of hallucinations generated in the models' responses.}
\label{fig3}
\end{figure*}

\subsubsection{Factual Uncertainty Questions}
In contrast to standard deterministic evaluation, factual uncertainty questions introduce scenarios that cannot be reliably resolved based on visual evidence alone. The core objective of this subset is to assess model behavior when facing genuine epistemic or aleatoric uncertainty, testing whether a model can recognize its own cognitive boundaries rather than fabricating ungrounded responses. To systematically cover different sources of ambiguity, we formulate three distinct types of uncertainty questions:

\paragraph{Information Deficiency Type} This category involves scenarios where the visual context lacks crucial, non-visual information that is absolutely necessary for definitive reasoning, such as personal identity, specific geographical location, or complex relational inference. To answer correctly, models must acknowledge the absence of sufficient evidence instead of guessing blindly. Typical examples include:
\begin{enumerate}
\item \emph{What is the name of the owner of this car in the image?}
\item \emph{In which city was this photo taken?}
\item \emph{Are the two people in the image sisters?}
\end{enumerate}

\paragraph{Subjective Judgment Type} These questions introduce queries involving aesthetic, emotional, or value-based evaluations that inherently lack objective, universal answers. They assess whether a model inappropriately projects a deterministic stance onto subjective human experiences. Exemplary formulations include:
\begin{enumerate}
\item \emph{Does the color scheme of the car in the image look nice for its owner?}
\item \emph{Is the skateboard in the image expensive for this skateboarder?}
\end{enumerate}

\paragraph{Visual Ambiguity Type} This class targets uncertainty arising directly from the limitations of visual indistinguishability within the image, such as fine-grained material composition, absolute physical scale, or object authenticity. These questions test the model's capacity to hedge its answers when visual patterns are inherently incomplete or deceptive. Examples of this type include:
\begin{enumerate}
\item \emph{Is the white clothing in the image made of pure cotton?}
\item \emph{How tall is the teddy bear in the image?}
\item \emph{Are the flowers on the table in the image real flowers?}
\end{enumerate}

\subsection{Uniqueness of CoSimUE}
As demonstrated by the comprehensive comparison in Table \ref{tab:1}, CoSimUE distinguishes itself from existing major benchmarks through several pioneering features designed to address current evaluation gaps.

First, in terms of dataset scale, CoSimUE contains 502 meticulously verified edited images, which represents the largest collection among comparable datasets dedicated to controlled visual perturbations. Second, regarding the perturbation methodology, CoSimUE uniquely generates diverse hallucination scenarios by leveraging both targeted textual modifications and random perturbations, whereas traditional benchmarks often rely on uniform manipulation techniques. Third, by focusing squarely on co-occurrence and similarity-driven hallucinations, our benchmark enables a remarkably fine-grained and isolated analysis of these two prevalent yet under-explored error types.

Finally, a core innovation of CoSimUE lies in its explicit treatment of model reliability under imperfect information. By introducing a specialized subset of 100 uncertainty questions that are posed directly on the perturbed images where no single, definitive answer exists, our dataset allows researchers to directly evaluate a model’s reasoning capabilities and hedge behavior under visual and semantic ambiguity. To the best of our knowledge, this makes CoSimUE the first benchmark explicitly designed to systematically assess and quantify uncertainty awareness in LVLMs.

\begin{table*}[t]
  \centering
  \caption{\textbf{Accuracy Leaderboard on CoSimUE with various LVLMs.} All the values are presented in \%, with a maximum score of 100\%.}
  \begin{threeparttable}
  \centering
  \footnotesize
  \begin{tabular}{@{}lccccccc@{}}
    \toprule
    Model & Param. & Total Pair Acc $\uparrow$ & Co-Pair Acc $\uparrow$ & Sim-Pair Acc $\uparrow$ & All Acc $\uparrow$ & Yes Bias & Pair Consist. $\downarrow$\\ 
    \midrule
    GPT-5 \cite{gpt5} & - & 72.66 & 71.01 & 75.86 & 85.64 & 55.47 & 25.59\\
    Claude Sonnet 4.5 \cite{claude4.5} & - & 60.16 & 58.88 & 62.64 & 78.91 & 57.81 & 36.91\\
    Gemini 2.5 Pro \cite{gemini2.5} & - & 67.38 & 64.79 & 72.41 & 83.30 & 61.82 & 31.45\\
    GLM-4.5 \cite{glm4.5} & - & \textbf{85.35} & \textbf{83.43} & \textbf{89.08} & \textbf{92.48} & 51.76 & \textbf{13.87}\\
    Grok 4 \cite{grok4} & - & 73.44 & 70.41 & 79.31 & 85.64 & 55.18 & 24.41\\
    Qwen3-VL \cite{qwen3} & - & 77.73 & 76.92 & 79.31 & 88.48 & 55.96 & 20.51\\
    \midrule
    LLaVA-1.6 \cite{llava1.6} & 13.4B & \textbf{65.63} & \textbf{62.43} & \textbf{71.84} & \textbf{82.62} & 65.04 & 33.98\\
    LLaVA-1.5 \cite{tuning} & 13.4B & 31.16 & 28.85 & 35.63 & 65.06 & 84.38 & 71.09\\
    InstructBLIP \cite{instructblip} & 12.3B & 29.10 & 23.96 & 39.08 & 63.18 & 69.73 & 67.97\\
    BLIP2-T5 \cite{blip2} & 12.2B & 31.84 & 32.25 & 31.03 & 65.04 & 33.50 & 65.82\\
    Qwen-VL \cite{Qwen-VL} & 9.7B & 58.01 & 55.03 & 63.79 & 75.00 & 43.75 & \textbf{32.62}\\
    MiniGPT5 \cite{minigpt5} & 9.5B & 27.73 & 27.22 & 28.74 & 54.69 & 39.36 & 52.73\\
    MiniGPT4 \cite{minigpt4} & 7.8B & 26.37 & 26.04 & 27.01 & 55.27 & 34.86 & 57.62\\
    mPLUG-Owl-v3 \cite{mplugowl3} & 8.1B & 44.73 & 44.67 & 44.83 & 72.36 & 76.66 & 55.27\\
    mPLUG-Owl-v2 \cite{mplugowl2} & 8.2B & 32.62 & 32.25 & 33.33 & 66.02 & 81.45 & 66.80\\
    mPLUG-Owl-v1 \cite{mplugowl} & 7.1B & 20.12 & 17.46 & 25.29 & 52.93 & 74.02 & 65.63\\
    \bottomrule
  \end{tabular}
  \begin{tablenotes}
    \tiny
    \item $\uparrow$ indicates higher is better; $\downarrow$ indicates lower is better.
  \end{tablenotes}
  \end{threeparttable}
  \label{tab:2}
\end{table*}

\begin{table*}[t]
  \centering
  \caption{\textbf{Uncertainty Score Leaderboard on CoSimUE evaluated under the multi-judge framework with various LVLMs.} The full scores for the three types of uncertainty are 50, 25, and 25 respectively, while the full score for all uncertainty is 100. IJ-STD denotes the Inter-Judge Standard Deviation of uncertainty scores.}
  \begin{threeparttable}
  \centering
  \footnotesize
  \begin{tabular}{@{}lcccccc@{}}
    \toprule
    Model & Parameter & I Uncertainty $\uparrow$ & II Uncertainty $\uparrow$ & III Uncertainty $\uparrow$ & All Uncertainty $\uparrow$ & IJ-STD $\downarrow$\\ 
    \midrule
    GPT-5 \cite{gpt5} & - & 18.27 & 11.94 & 8.36 & 38.57 & 0.0126\\
    Claude Sonnet 4.5 \cite{claude4.5} & - & \textbf{37.18} & \textbf{17.63} & 15.22 & \textbf{70.03} & 0.0113\\
    Gemini 2.5 Pro \cite{gemini2.5} & - & 26.41 & 16.12 & 11.73 & 54.26 & 0.0123\\
    GLM-4.5 \cite{glm4.5} & - & 27.34 & 8.92 & 12.48 & 48.74 & 0.0119\\
    Grok 4 \cite{grok4} & - & 26.72 & 15.88 & \textbf{16.03} & 58.63 & 0.0113\\
    Qwen3-VL \cite{qwen3} & - & 25.91 & 9.34 & 9.68 & 44.93 & 0.0109\\  
    \midrule
    LLaVA-1.6 \cite{llava1.6} & 13.4B & \textbf{30.62} & \textbf{16.48} & \textbf{12.93} & \textbf{60.03} & 0.0101\\
    LLaVA-1.5 \cite{tuning} & 13.4B & 5.37 & 6.52 & 3.74 & 15.63 & 0.0043\\
    InstructBLIP \cite{instructblip} & 12.3B & 8.11 & 1.73 & 1.44 & 11.28 & 0.0031\\
    BLIP2-T5 \cite{blip2} & 12.2B & 8.02 & 0.36 & 0.58 & 8.96 & 0.0029\\
    Qwen-VL \cite{Qwen-VL} & 9.7B & 1.24 & 0.67 & 2.58 & 4.49 & 0.0021\\
    MiniGPT5 \cite{minigpt5} & 9.5B & 18.09 & 9.42 & 8.17 & 35.68 & 0.0079\\
    MiniGPT4 \cite{minigpt4} & 7.8B & 23.94 & 11.27 & 10.03 & 45.24 & 0.0083\\
    mPLUG-Owl-v3 \cite{mplugowl3} & 8.1B & 12.61 & 6.73 & 4.21 & 23.55 & 0.0058\\
    mPLUG-Owl-v2 \cite{mplugowl2} & 8.2B & 7.42 & 5.11 & 0.82 & 13.35 & 0.0046\\
    mPLUG-Owl-v1 \cite{mplugowl} & 7.1B & 6.93 & 9.84 & 1.62 & 18.39 & 0.0051\\
    \bottomrule
  \end{tabular}
  \begin{tablenotes}
      \tiny
      \item $\uparrow$ indicates higher is better; $\downarrow$ indicates lower is better.
  \end{tablenotes}
  \end{threeparttable}
  \label{tab:3}
\end{table*}

\section{Evaluation Protocol}
To ensure a fair and consistent comparison across various models, we establish a standardized evaluation protocol for utilizing our benchmark, as detailed below.

\subsection{Data Usage}

All experiments are conducted exclusively on the CoSimUE benchmark under a strict zero-shot setting, without any additional training data, external datasets, or task-specific fine-tuning. Only the provided original and perturbed image-question pairs are utilized for inference. This isolation ensures that performance discrepancies primarily reflect architectural design choices rather than variations in data scale or supervision effects.

\subsection{Certainty Evaluation Metrics}
To rigorously evaluate the model's resistance against co-occurrence and similarity hallucinations, we establish a specialized suite of evaluation metrics tailored for factual certainty questions. This comprehensive evaluation framework consists of four complementary metrics, namely Pair Accuracy, All Accuracy, Yes Bias, and Pair Consistency.

\subsubsection{Pair Accuracy}
Pair Accuracy serves as a stringent joint metric that quantifies whether the model can correctly predict the targeted object-level relationships across both the original image and its corresponding hallucinated counterpart simultaneously. We report Total, Co-occurrence, and Similarity Pair Accuracy based on different perturbation subsets, which is formally defined as:
\begin{equation}
\text{Accuracy}_{\mathcal{C}} = \frac{\sum_{i=1}^{N_{\mathcal{C}}} \mathbb{I} \left( \hat{y}_i^{\text{orig}} = y_i^{\text{orig}} \land \hat{y}_i^{\text{hallu}} = y_i^{\text{hallu}} \right)}{N_{\mathcal{C}}}
\label{eq:2}
\end{equation}
\(\hat{y}_i^{\text{orig}}\) and \(\hat{y}_i^{\text{hallu}}\) are the model’s predictions for the original and hallucinated images. \(y_i^{\text{orig}}\) and \(y_i^{\text{hallu}}\) denote their corresponding ground-truth answers. \(\mathcal{C}\) represents the category (\(\mathcal{T}\) for total, \(\mathcal{L}\) for co-occurrence, \(\mathcal{S}\) for similarity).

\subsubsection{All Accuracy} All Accuracy measures the global accuracy of the model by treating all test instances independently across the combined pool of original and hallucinated images, providing an overall reflection of factual correctness:
\begin{equation}
aAcc = \frac{\sum_{i=1}^{N} \mathbb{I} \left( \hat{y}_i^{\text{orig}} = y_i^{\text{orig}} \right) + \sum_{i=1}^{N} \mathbb{I} \left( \hat{y}_i^{\text{hallu}} = y_i^{\text{hallu}} \right)}{2N}
\label{eq:4}
\end{equation}

\subsubsection{Yes Bias} Yes Bias quantifies the fundamental tendency of a model to output affirmative answers under a perfectly balanced Yes/No distribution enforced by our label-flip setup (50\%/50\%), thereby capturing the model's global over-affirmation or sycophancy behavior:
\begin{equation}
B_y= \frac{\sum_{i=1}^{N} \mathbb{I} \left( \hat{y}_i^{\text{all}} = \text{"yes"} \right)}{2N}
\label{eq:5}
\end{equation}

\subsubsection{Pair Consistency} Pair Consistency evaluates whether the model's predictions remain identical when presented with the original and the hallucinated image within the same pair. Since each evaluation pair enforces a deterministic ground-truth label flip from Yes to No through controlled object-level perturbations, a high value of \(P_c\) coupled with poor accuracy directly exposes the model's tendency to rely blindly on persistent contextual priors or background features:
\begin{equation}
P_c = \frac{\sum_{i=1}^{N} \mathbb{I} \left( \hat{y}_i^{\text{orig}} = \hat{y}_i^{\text{hallu}} \right)}{N}
\label{eq:6}
\end{equation}

\subsection{Uncertainty Evaluation Metrics}
In this section, we introduce the Uncertainty Evaluation Metrics, which assess the degree of uncertainty in the model's responses. These metrics are designed to quantify how effectively the model handles ambiguous or uncertain scenarios.

\subsubsection{Uncertainty Scoring Method}
To systematically assess response uncertainty, we adopt a multi-judge framework where multiple advanced LLMs independently score each generated answer on a continuous scale from 0 to 1. Given the 100 uncertainty questions in our benchmark, the summed score corresponds to a normalized 100-point uncertainty scale. The evaluation prompt for the judges is formulated as follows:

\emph{You will evaluate whether the text I provide is a definite statement or an uncertain response. A definite statement provides clear and conclusive information. An uncertain response contains ambiguity or lack of clarity. If the text is a definite statement, reply with 0. If the text is uncertain, reply with a value between 0 and 1. The closer the response is to 1, the more uncertain the text is. }

\subsubsection{Uncertainty Score Calculation}
The final uncertainty score for each model response is obtained by aggregating and averaging the evaluation scores assigned by our multi-judge panel, which consists of state-of-the-art large language models including GPT-5, Claude Sonnet 4.5, Gemini 2.5 Pro, GLM-4.5, Grok 4, and Qwen3-VL. By leveraging the collective consensus of these diverse architectures, we minimize the subjective bias inherent in any single evaluation model. he comprehensive All Uncertainty Score \(U\) is defined as:
\begin{equation}
U = \sum_{i=1}^{N} \frac{1}{K} \sum_{j=1}^{K} M_j(r_i)
\label{eq:7}
\end{equation}
where \(N\) represents the total number of evaluated responses, \(K\) denotes the total number of judges in the panel, and \(M_j(r_i)\) signifies the continuous uncertainty score assigned by the \(j\)-th judge to the \(i\)-th response \(r_i\).

\subsubsection{Inter-Judge Standard Deviation} To formally quantify the reliability and consensus of the multi-judge framework, we introduce the Inter-Judge Standard Deviation. This metric measures the dispersion of the assigned uncertainty scores across the different judges for each individual response, reflecting the overall level of inter-judge agreement throughout the dataset. The standard deviation \(\sigma\) is calculated as:
\begin{equation}
\sigma = \frac{1}{N} \sum_{i=1}^{N} \sqrt{\frac{1}{K} \sum_{j=1}^{K} \left(M_j(r_i) - \bar{M}(r_i)\right)^2}
\label{eq:8}
\end{equation}
where \(\bar{M}(r_i)\) denotes the mean uncertainty score of the response \(r_i\) across all \(K\) judges. Statistically, a lower value of \(\sigma\) indicates a stronger cross-model agreement among the judges, thereby validating the robustness and objectivity of the automated scoring pipeline.

\subsection{Inference Settings}
All baseline models are evaluated under a unified inference configuration to guarantee a rigorous and unbiased comparison. Specifically, we adopt a deterministic decoding strategy by setting the temperature to 0 and disabling beam search (num\_beams = 1). Additionally, no nucleus sampling top\_p is applied, and the maximum number of generated tokens is strictly limited to 512.

\section{Experiments and Insights}
\begin{figure*}[t]
  \centering
  \subfloat[Overall performance.\label{fig:short-a}]{
    \includegraphics[width=0.45\textwidth]{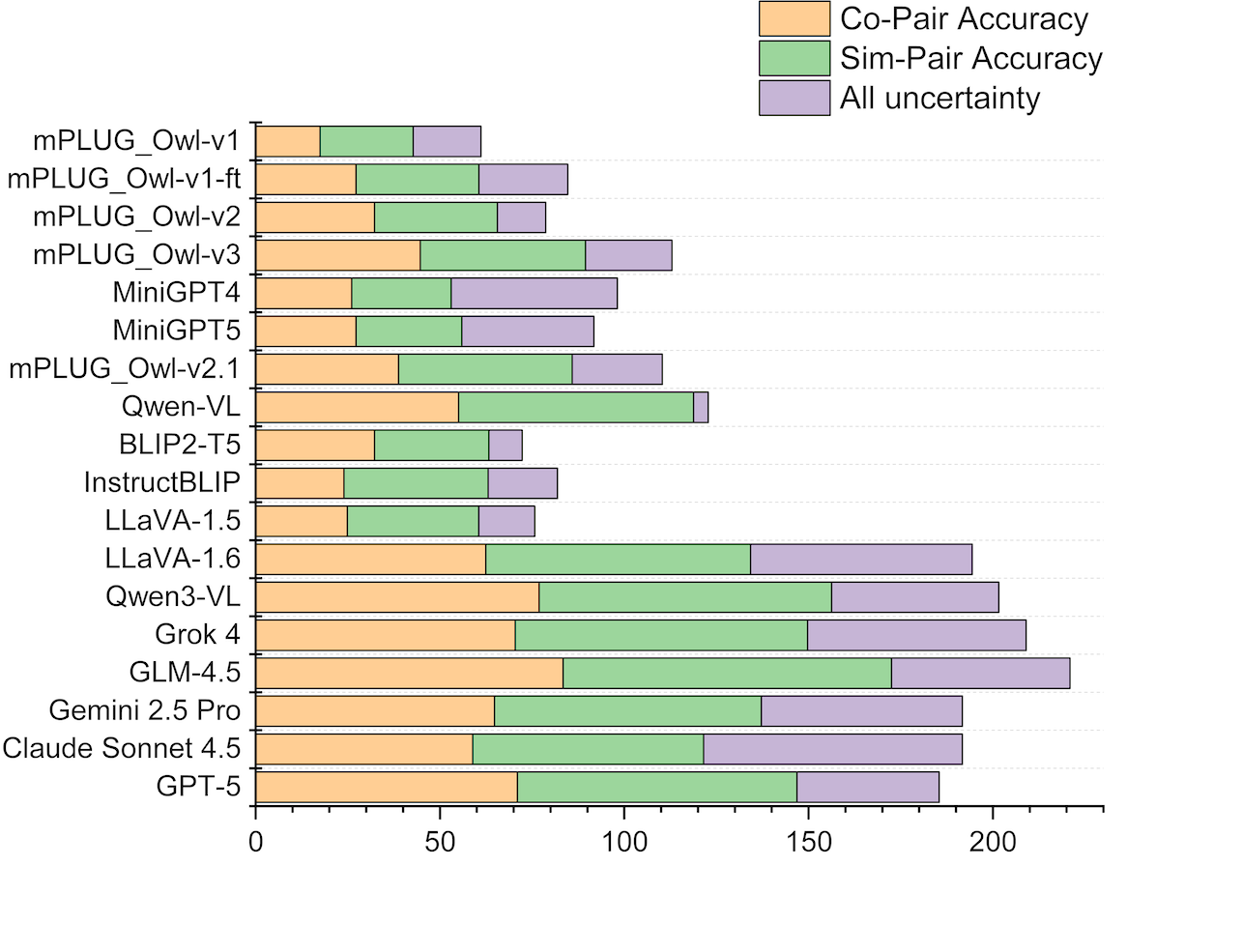}
  }
  \hfil
  \subfloat[Radar chart.\label{fig:short-b}]{
    \includegraphics[width=0.45\textwidth]{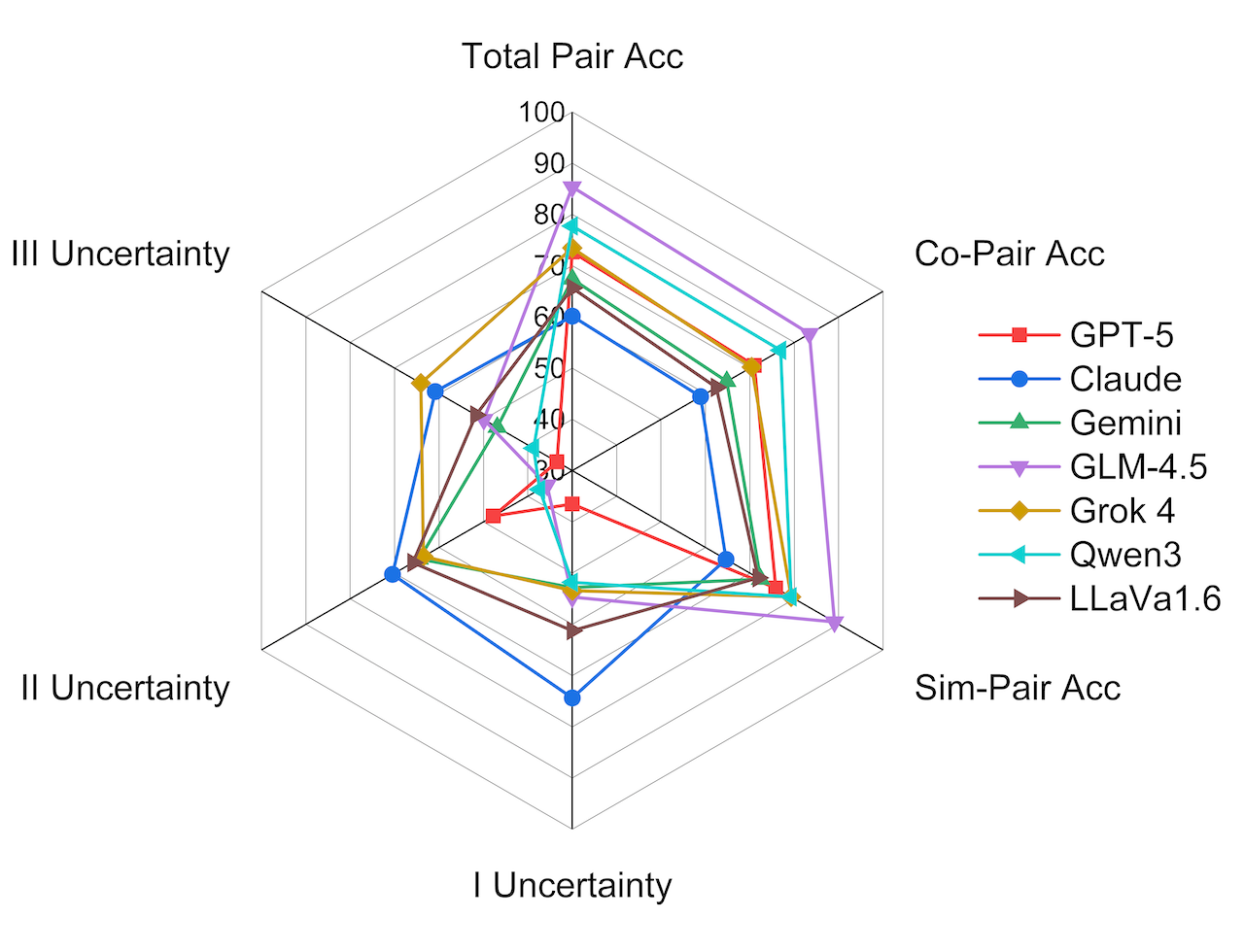}
  }
  \caption{Double-column width hallucination analysis.}
  \label{fig4}
\end{figure*}

Based on the proposed CoSimUE benchmark, we conduct extensive empirical experiments across a diverse set of prominent LVLM architectures. These evaluated models span a wide spectrum of parameter scales, core design choices, and training paradigms, encompassing both open-source frameworks and representative proprietary models to ensure a comprehensive evaluation. The quantitative evaluation results are comprehensively documented from Table \ref{tab:2} to Table \ref{tab:8}, while the global performance comparisons and trade-offs are illustrated in Fig. \ref{fig4}. Specifically, our multi-dimensional study is systematically organized around three primary pillars, namely the Linguistic Foundation, the Visual Representation, and the Semantic Alignment. Within these pillars, we thoroughly investigate 7 distinct architectural design aspects. Furthermore, we also conduct series of pairwise fusion experiments to carefully analyze the cross-dimensional interactions and synergistic effects on different hallucination categories.

\subsection{Linguistic Foundation}

\subsubsection{Language model’s parameter scale} As demonstrated by the empirical evaluations documented in Table \ref{tab:2} and Table \ref{tab:3}, proprietary closed-source models generally exhibit a consistent performance advantage over smaller open-source models across both co-occurrence and similarity hallucination categories. However, a distinct counter-trend is observed when evaluating uncertainty hallucinations. In this non-deterministic scenario, several leading closed-source models deliver a lower accuracy, underperforming in comparison to the open-source LLaVA-1.6 framework. Furthermore, looking closer at the structural impact of model capacity as illustrated in Table \ref{tab:4}, our data indicates that the benefits of parameter scaling are highly concentrated, primarily contributing to the mitigation of co-occurrence errors.

\begin{table}[ht]
  \caption{Comparison of different language model parameter scales on CoSimUE.}
  \centering
  \tiny
  \begin{tabular}{@{}lllllccccc@{}}
  \toprule
  Model & Language Model & Co-Pair & Sim-Pair & All Uncertainty\\ 
  \midrule
  LLaVA-1.6 & Vicuna-7B & 60.95 & 71.84  & 56.64 \\
  LLaVA-1.6 & Vicuna-13B & 62.43(+1.48) & 71.84(+0.00) & 60.03(+3.39)\\
  \midrule
  BLIP2-T5 & Flant5xl & 30.91 & 43.28 & 4.68\\
  BLIP2-T5 & Flant5xxl & 32.25(+1.34) & 31.03(-12.25) & 8.96(+4.28)\\
  \midrule
  InstructBLIP & Flant5xl & 9.17 & 27.52 & 8.17\\
  InstructBLIP & Flant5xxl & 23.96(+14.79) & 39.08(+11.56) & 11.28(+3.11)\\
  \bottomrule
  \end{tabular}
  \label{tab:4}
\end{table}

\insight{Parameter expansion slightly enhances linguistic reasoning and co-occurrence control, and generally offers limited benefit for uncertainty awareness.}

\subsubsection{Network structure} As summarized in Table \ref{tab:2}, GLM-4.5 \cite{glm4.5} achieves the strongest overall performance when mitigating both co-occurrence and similarity hallucinations. From an architectural perspective, this framework introduces distinct modifications compared with conventional vision-language models. Specifically, GLM-4.5 adopts a structurally deeper and narrower network configuration, implements a significantly higher number of attention heads within its layers, and incorporates a dedicated visual localization module into its processing pipeline.

\begin{table}[ht]
  \caption{Comparison of different language models of the same scale on CoSimUE.}
  \centering
  \tiny
  \begin{tabular}{@{}lllllccccc@{}}
  \toprule
  Model & Language Model & Co-Pair & Sim-Pair & All Uncertainty \\ 
  \midrule
  LLaVA-1.5 & LLaMA-7B & 32.67 & 46.31 & 9.77 \\
  LLaVA-1.5 & Vicuna-7B & 34.91(+2.24) & 47.70(+1.39) & 11.83(+2.06)\\
  \midrule
  LLaVA-1.6 & Vicuna-7B & 60.95 & 71.84 & 56.65\\
  LLaVA-1.6 & Mistral-7B & 62.57(+1.62) & 72.19(+0.35) & 60.34(+3.69)\\
  \midrule
  MiniGPT4 & LLaMA-7B & 24.37 & 26.63 & 43.46\\
  MiniGPT4 & Vicuna-7B & 26.04(+1.67) & 27.01(+0.38) & 45.24(+1.78)\\
  \bottomrule
  \end{tabular}
  \label{tab:5}
\end{table}

\insight{Increasing architectural depth and attention diversity, as well as employing visual localization modules, is more effective for reducing co-occurrence and similarity hallucinations than merely expanding model width.}

\subsubsection{Training mechanism} As demonstrated by the empirical comparative results presented in Table \ref{tab:5}, replacing the baseline language model, which is trained based on the standard LLaMA-7B architecture, consistently yields a noticeable reduction in overall hallucinations. Among the alternatives evaluated, Vicuna \cite{vicuna} utilizes a large-scale instruction tuning paradigm built upon the foundation of LLaMA \cite{llama}. Meanwhile, Mistral \cite{mistral} incorporates a more refined alignment process along with a specialized sliding-window attention mechanism. Finally, Claude Sonnet 4.5 \cite{claude4.5}, which adopts a structured Constitutional AI framework for its underlying optimization, achieves the highest uncertainty score across the entire evaluation suite.

\vspace{30pt}
\insight{Advanced training strategies such as large-scale instruction tuning and refined attention mechanisms play a critical role in reducing hallucinations, particularly by strengthening linguistic grounding and improving uncertainty awareness beyond what scaling alone can achieve.}

\subsection{Visual Representation}

\subsubsection{Vision encoder} As demonstrated by the empirical evaluation documented in Table \ref{tab:6}, upgrading the foundational visual encoder from the standard CLIP-L \cite{clip} configuration to the enhanced CLIP-L/336px variant yields a consistent improvement in model performance. This positive trend is particularly prominent within the similarity hallucination category. Structurally, this modification involves a direct transition in the input specification, where the model utilizes a significantly higher input resolution that delivers a finer, more detailed visual feature representation.

\begin{table}[ht]
  \caption{Comparison of different visual encoders and input processing strategies on CoSimUE.}
  \centering
  \tiny
  \begin{tabular}{@{}lllllccccc@{}}
  \toprule
  Model & Vision Encoder & Co-Pair & Sim-Pair & All Uncertainty\\ 
  \midrule
  BLIP2-T5 & ViT-L & 32.25 & 31.03 & 8.96\\
  InstructBLIP & ViT-L* & 23.96(-8.29) & 39.08(+8.05) & 11.28(+2.32)\\
  \midrule
  LLaVA-1.3-7B & CLIP-L & 25.93 & 31.39 & 10.76\\
  LLaVA-1.3-7B & CLIP-L/336px & 27.69(+1.76) & 34.42(+3.03) & 9.11(-1.65)\\
  \midrule
  LLaVA-1.5-7B & CLIP-L & 33.71 & 44.58 & 12.37\\
  LLaVA-1.5-7B & CLIP-L/336px & 34.91(+1.20) & 47.70(+3.12) & 11.83(-0.54)\\
  \bottomrule
  \end{tabular}
  \label{tab:6}
\end{table}

\subsubsection{Processing strategy for visual inputs} As demonstrated by the empirical comparisons in the evaluation, the architectural transition from the baseline BLIP2 \cite{blip2} framework to InstructBLIP \cite{instructblip} introduces both the integrated Q-former module and an enhanced instruction-following tuning paradigm. This specific modification yields a highly distinct, bifurcated impact on different error categories. On one hand, the architectural shift leads to a noticeable reduction in similarity hallucinations. On the other hand, this improvement is accompanied by a simultaneous increase in co-occurrence hallucinations, presenting an intriguing trade-off across the two dimensions.

\insight{Enhancing visual representation quality through higher resolution encoders and refined visual processing modules significantly reduces similarity hallucinations by improving object-level discrimination, though tighter linguistic coupling may slightly increase co-occurrence bias.}
\subsection{Semantic Alignment}

\subsubsection{Alignment degree} As demonstrated by the empirical results presented in Table \ref{tab:7}, mPLUG-Owl-v1 \cite{mplugowl} exhibits a clear and continuous performance trajectory across different training paradigms. Specifically, when tracking the model's development from the initial pre-training stage to LoRA tuning, and ultimately to full parameter fine-tuning, there is a progressive and steady improvement in its capabilities. This upward trend is consistently observed in both the mitigation of co-occurrence hallucinations and the enhancement of uncertainty performance.
\begin{table}[ht]
  \caption{Comparison of alignment degrees on CoSimUE.}
  \centering
  \tiny
  \begin{tabular}{@{}lllllcccccc@{}}
  \toprule
  Model & Schedule & Co-Pair & Sim-Pair & All Uncertainty \\ 
  \midrule
  LLaVA-1.5-13B & LoRA & 22.15 & 34.21 & 14.82\\ 
  LLaVA-1.5-13B & Full & 28.85(+6.70) & 35.63(+1.42) & 15.63(+0.81)\\
  \midrule
  mPLUG-Owl-v1-pt & Pretrain & 6.80 & 17.75 & 11.81\\
  mPLUG-Owl-v1 & LoRA & 17.46(+10.66) & 25.29(+7.54) & 18.39(+6.58)\\
  mPLUG-Owl-v1-ft & Full & 27.22(+20.42) & 33.33(+15.58) & 24.12(+12.31)\\
  \bottomrule
  \end{tabular}
  \label{tab:7}
\end{table}

\subsubsection{Data quality} As demonstrated by the quantitative results, each evaluated model exhibits distinct behavior patterns that correspond directly with their respective configurations and utilized fine-tuning datasets. Specifically, Qwen-VL \cite{Qwen-VL} demonstrates an exceptionally high level of prediction accuracy, which is accompanied by notably low uncertainty scores across our evaluation suite. In sharp contrast, the MiniGPT series \cite{minigpt4, minigpt5} records significantly higher uncertainty scores. In terms of data composition, the training pipeline of the MiniGPT series incorporates a highly curated, high-quality image--text dataset, defining a different empirical profile compared to the Qwen framework.

\insight{Increasing the proportion of model parameters involved in cross-modal alignment, together with high-quality and well-curated data, enhances factual consistency and uncertainty awareness, suggesting that alignment capacity and data quality substantially influence model reliability.}

\begin{table}[ht]
  \caption{Comparison of different design dimension combinations on CoSimUE.}
  \centering
  \tiny
  \begin{tabular}{@{}lllllcccccc@{}}
  \toprule
  Model & Design Comb & Co-Pair & Sim-Pair & All Uncertainty \\ 
  \midrule
  mPLUG-Owl-v2 & - & 32.25 & 33.33 & 13.35 \\
  mPLUG-Owl-v2.1 & LF+VR & 38.76(+6.51) & 47.13(+13.80) & 24.57(+11.22)\\
  \midrule
  LLaVA-1.3-7B & - & 21.85 & 29.76 & 5.29\\
  LLaVA-1.3-7B* & LF+SA & 27.69(+5.84) & 34.42(+4.66) & 9.11(+3.82)\\
  \midrule
  LLaVA-1.5-7B & - & 34.91 & 47.70 & 11.83\\
  LLaVA-1.6-7B & VR+SA & 60.95(+26.04) & 71.84(+24.14) & 56.65(+44.82)\\
  \bottomrule
  \end{tabular}
  \label{tab:8}
\end{table}

\subsection{Cross-Dimensional Analysis}
As demonstrated by the quantitative results documented in Table \ref{tab:8}, the architectural transition in mPLUG-Owl2, which replaces the baseline ViT-L visual encoder with ViT-G and substitutes LLaMA with the Qwen language foundation, leads to a distinct performance improvement specifically within the similarity hallucination category. Meanwhile, evaluating the configuration of LLaVA-1.3 \cite{llava} reveals that the integration of the Vicuna language model coupled with a full parameter fine-tuning paradigm results in a noticeable reduction in co-occurrence hallucinations. Furthermore, the upgraded LLaVA-1.6 framework incorporates both higher-resolution visual inputs and a significantly higher-quality instruction dataset, yielding comprehensive performance gains across all evaluated hallucination types. Ultimately, the empirical data shows that the joint combination of advanced Visual Representation and robust Semantic Alignment correlates with the largest overall reduction in hallucinations across the entire benchmark suite.

\vspace{30pt}
\insight{ Visual input remains as a weaker component in current LVLMs, and jointly improving visual fidelity and alignment quality is key to building more reliable, hallucination-resilient models.}

\section{Conclusion}
In this paper, we present the first systematic investigation into how architecture-level design choices influence the multi-faceted nature of hallucinations in LVLMs. By decomposing the architectural design space into Linguistic Foundation, Visual Representation, and Semantic Alignment, and evaluating them across Co-occurrence, Similarity, and Uncertainty dimensions via our CoSimUE benchmark, we uncover several critical design–reliability relationships. Specifically, our empirical evaluation demonstrates that: (1) LLMs offer surprisingly limited gains for resolving uncertainty; (2) stronger visual representations primarily serve to reduce similarity-based hallucinations; (3) improved semantic alignment effectively mitigates co-occurrence errors by breaking incorrect statistical associations; and (4) combining advanced visual representation with robust semantic alignment achieves the most comprehensive and effective reduction in overall hallucinations. These insights suggest that future trustworthy multimodal systems must prioritize synergistic architectural design over singular linguistic scaling.

\bibliographystyle{IEEEtran}
\bibliography{references}

\vfill

\end{document}